\documentclass{article}

 \usepackage[dblblindworkshop, final]{neurips_2025}

\workshoptitle{EurIPS 2025 Workshop UPLB}

\usepackage[utf8]{inputenc} 
\usepackage[T1]{fontenc}    
\usepackage{hyperref}       
\usepackage{url}            
\usepackage{booktabs}       
\usepackage{amsfonts}       
\usepackage{nicefrac}       
\usepackage{microtype}      
\usepackage{xcolor}         

\usepackage{linguex}

\usepackage{tabularx}
\usepackage{booktabs}

\usepackage{amssymb}  
\usepackage{amsmath}

\usepackage{graphicx}

\usepackage{float}

\usepackage{caption}
\usepackage{tikz}
\usepackage{pgfplots}
\usepgfplotslibrary{groupplots}
\pgfplotsset{compat=1.18}

\title{Uncovering Implicit Bias in LLM Mathematical Reasoning with Concept Learning}

\author{%
  Leroy Z. Wang  \\
  Independent\\
  \texttt{wang.leroy \_AT\_ outlook \_DOT\_ com}
}

\begin{document}

\maketitle

\begin{abstract}
We introduce a new framework of concept learning tasks that helps uncover implicit biases in large language models. Using in-context concept learning experiments, we found that language models can have a bias toward upward monotonicity; such bias is less apparent when the model is tested by direct prompting without concept learning components. This demonstrates that in-context concept learning can be an effective way to discover hidden biases in large language models.
\end{abstract}

\section{Introduction}

As Large Language Models (LLMs) become important components of various Natural Language Processing (NLP) systems, there is an increased scrutiny on the potential biases models 
may have before deployment. Researchers have developed benchmarks and de-biasing methods that help to detect and mitigate biases in LLMs \citep{bias-01-wan-etal-2023-kelly, bias-03-Hofmann2024AIGC, bias-02-shirafuji-etal-2025-bias}; however, recent studies show that models that appear unbiased on standard benchmarks can still have implicit biases that are hard to detect \citep{implicit-bias-pnas, implicit-bias-02-tan-lee-2025-unmasking}. 

To help uncover such implicit biases, we propose a new method inspired by the literature on human concept learning \citep{feldmanMinimizationBooleanComplexity2000, goodmanConceptsProbabilisticLanguage2015, piantadosiLogicalPrimitivesThought2016}. In this paper, we use in-context concept learning to show how LLMs can have an implicit bias toward certain mathematical concepts. 

Consider the prompt in \ref{ex:alice}.  In the first four lines, we see two labeled examples of an unknown mathematical concept, expressed by ``the desired quantity''.  The last two lines ask the model to label a new example. By repeating this process for a variety of scenarios and concepts, we can study whether the model achieves better accuracy with mathematical concepts that have certain properties.

\ex. There are 10 boxes. Alice has 5 of the 10 boxes. Does Alice have the \textit{desired quantity} of the boxes? \textbf{No}.
\\
There are 15 boxes. Alice has 8 of the 15 boxes. Does Alice have the \textit{desired quantity} of the boxes? \textbf{Yes}.
\\
There are 16 boxes.  Alice has 9 of the 16 boxes.  Does Alice have the \textit{desired quantity} of the boxes? \_\_\_ \label{ex:alice}

We will use the idea of monotonicity from the study of semantics to formulate our experiments.  Here, we give an informal introduction to semantic monotonicity in quantifiers, which are semantic objects that describe the relation between two sets of objects in a discourse \citep{quantifiers-Barwise1981GeneralizedQA}. For example, \textit{more than half} is a quantifier in English, which can be a possible meaning for the unknown concept in \ref{ex:alice}, since we can substitute \textit{more than half} for \textit{the desired quantity} in the first two examples while satisfying their corresponding labels. A quantifier is upward monotone if inferences from subsets to supersets are valid \footnote{In this paper, we adopt a simplified definition of quantifier monotonicity, and only consider the monotonicity of the second argument taken by the quantifier. For a more rigorous treatment, see \citet{icard-iii-moss-2014-monotonicity}.}, and it is downward monotone if inferences from supersets to subsets are valid
\citep{icard-iii-moss-2014-monotonicity, monotone-quant-01-Carcassi2021MonotoneQE}. See table \ref{monotone-examples} for some concrete examples.

\begin{table}[ht]
\centering

\begin{tabular}{ ll } 
\toprule

\textit{\large Upward monotone}  & \; \\

more than $n$ & More than 5 boxes are in Berlin. $\Rightarrow$ More than 5 boxes are in Germany.\\ 

some & Some boxes are in Berlin. $\Rightarrow$ Some boxes are in Germany.\\

\midrule

\textit{\large Downward monotone}  & \; \\

less than $n$ & Less than 5 boxes are in Germany. $\Rightarrow$ Less than 5 boxes are in Berlin.\\ 

no & No boxes are in Germany. $\Rightarrow$ No boxes are in Berlin.\\

\bottomrule
\end{tabular}

\caption{Monotonicity in quantifiers.}
\label{monotone-examples}
\end{table}

We test three different LLMs on a range of mathematical concepts that differ in semantic monotonicity, and find that models tend to have greater success with upward monotone concepts. We then performed the same set of experiments with explicit semantics, that is, instead of hiding the meaning of the concept using the phrase ``the desired quantity'', we replaced the phrase with the description of the concept's meaning in plain English\footnote{e.g. the question in the prompt can be ``Does Alice have \textit{more than 1/2}  of the boxes?''}. We find that the bias toward upward monotonicity becomes less noticeable. This suggests that concept learning can be used to uncover implicit biases in mathematical reasoning that are hard to detect with standard methods.

\section{Related Work}

\citet{min-etal-2022-rethinking, in-context-learn-01-Wei2022EmergentAO, in-context-learn-02-Kojima2022LargeLM}, among many others, have studied how and why in-context learning in LLMs works.
\citet{implicit-bias-pnas} demonstrated that LLMs that appear unbiased on common social bias (such as gender and race) benchmarks can still have implicit biases in their decision-making process. Our work follows a similar theme, but instead of social biases, we study implicit biases in LLMs' mathematical reasoning process. \citet{monotonicity-processing-load-Geurts2005} showed that humans generally achieve higher accuracy for upward monotone quantifiers (vs. downward monotone) in quantifier inference tasks, which suggests downward monotone quantifiers can be more cognitively complex for humans. \citet{monotone-npi-license} studied how LLMs use semantic monotonicity to process negative polarity items in English. And \citet{minimization-boolean-complexity-incontext} showed that similar to humans, LLMs can have a simplicity bias in concept learning that favors logically simpler concepts. 

\section{Methodology}

\subsection{Concept selection}

For concepts that are upward monotone, we use the quantifier ``more than $p$'', where $p$ represents a proportion of the total number of items. For downward monotone concepts, we use the quantifier ``less than $p$''. We use the following values for $p$ in all experiments:
\begin{verbatim}
    p = {1/10, 2/10, 3/10, 4/10, 5/10, 6/10, 7/10, 8/10, 9/10}.
\end{verbatim}

See Appendix \ref{additional-concepts} for additional proportional concepts and cardinal concepts (e.g. ``more/less than $c$'' where $c$ is a cardinal number) results.

\subsection{Prompt generation}

In each prompt, we include 20 labeled examples (10 with positive labels and 10 with negative labels) in a random order. Each example is generated using the template in \ref{ex:template}, where the underlined slots denote the linguistic items that may vary between examples. See \ref{example-prompt} in Appendix for an example prompt used in the experiments.

\ex. There are  $\underset{\text{TOTAL}}{\underline{\hspace{1cm}}}$ \, $\underset{\text{OBJ}}{\underline{\hspace{1cm}}}$. \;  Alice  has  \, $\underset{\text{NUM}}{\underline{\hspace{1cm}}}$ of the \, $\underset{\text{TOTAL}}{\underline{\hspace{1cm}}}$ \, $\underset{\text{OBJ}}{\underline{\hspace{1cm}}}$.\\ 
\\
Does Alice have the desired quantity of the $\underset{\text{OBJ}}{\underline{\hspace{1cm}}}$? \, $\underset{\text{YES/NO}}{\underline{\hspace{1cm}}}$.   \label{ex:template}

We iterate through all meaningful\footnote{e.g. Alice cannot have more items than the total number of items, or have less than 0 items. We also require the total number of items to be between 5 and 100 inclusive.} numerical ranges for both the number of total objects, and the number of objects Alice has, and generate an example for each combination. For each prompt, we randomly sample without replacement 10 positive examples and 10 negative examples from the sets of all possible positive examples and negative examples respectively. An unseen example is used as the question at the end.

For the set of objects in the prompt, we first took all the countable nouns from The Oxford 3000 word list, which is a (public-domain) list of 3000 most important words to learn in English, chosen by language experts at Oxford Dictionaries.
\footnote{\url{https://www.oxfordlearnersdictionaries.com/us/wordlist/american_english/oxford3000/}} We then ranked the nouns by their unigram frequency in OLMo-2-32B's training corpus, and chose the top 100 most frequent nouns as the set of objects in the prompt. For each example, the object is randomly sampled from the set of 100 nouns.


\subsection{Evaluation}
To determine the model's response to a prompt, we compare the probabilities (computed by the model) of the prompt with a positive response appended at the end, with the same prompt with a negative response appended. In other words, if $P(prompt  + \text{``Yes''}) > P(prompt  + \text{``No''})$, then we consider the model has given a positive response to the prompt.
For each concept, there are 250 prompts with positive true labels and 250 prompts with negative true labels. As a baseline (random), a model that answers every prompt by flipping a fair coin would achieve $50\%$ accuracy.

\subsection{Models}

We ran experiments on two LLM families that are fully open-source: OLMo 2 \citep{OLMo20242O2} from Allen Institute for AI (Ai2) and K2 \citep{k2-llm} from LLM360.  OLMo 2 is a family of fully open language models that achieve competitive performance on various benchmarks when compared to other open-weight models. K2-65B is a fully open model that outperforms Llama-2-70B \citep{Touvron2023Llama2O} on most benchmarks while using less compute. We chose models that are fully open (open model weights, model checkpoints, and exact training data) so that our experiments will have better reproducibility; having full access to the model training data will also make explaining model behaviors from training data possible in future work.

\section{Results}

\begin{figure}[H]
    \centering
\resizebox{\textwidth}{!}{

\begin{tikzpicture} 
\begin{groupplot}[
  group style={
    group size=3 by 2,
    horizontal sep=0.5cm,
    vertical sep=1cm
  },
  width=6cm,
  height=4cm,
  xlabel={ },
  ylabel={Accuracy \%},
  ymin=0, ymax=100,
  legend style={at={(2, 2)},anchor=north,legend columns=-1},
  symbolic x coords={$\frac{1}{10}$, $\frac{2}{10}$, $\frac{3}{10}$, $\frac{4}{10}$, $\frac{5}{10}$, $\frac{6}{10}$, $\frac{7}{10}$, $\frac{8}{10}$, $\frac{9}{10}$ },
  xtick=data,
  grid=both,
]

\nextgroupplot[title={OLMo-2-13B-instruct}, xticklabels={}, ylabel={}]
\addplot+[mark=square*, orange, line width=2pt,   
nodes near coords,
  point meta=y,
  every node near coord/.append style={font=\scriptsize, yshift=2pt}] coordinates {
  ($\frac{1}{10}$, 81.2) ($\frac{2}{10}$, 76) ($\frac{3}{10}$, 77) ($\frac{4}{10}$, 76.8) ($\frac{5}{10}$, 79) ($\frac{6}{10}$, 75.6) ($\frac{7}{10}$, 77) ($\frac{8}{10}$, 78) ($\frac{9}{10}$, 79) 
};
\addlegendentry{[more than $p$]}
\addplot+[mark=*, teal, line width=1pt,
nodes near coords,
  point meta=y,
  every node near coord/.append style={font=\scriptsize, yshift=2pt}] coordinates {
  ($\frac{1}{10}$, 51) ($\frac{2}{10}$, 47.6 ) ($\frac{3}{10}$, 41.2 ) ($\frac{4}{10}$, 39) ($\frac{5}{10}$, 42.8 ) ($\frac{6}{10}$, 38.4 ) ($\frac{7}{10}$, 37 ) ($\frac{8}{10}$, 36.6 ) ($\frac{9}{10}$, 36.8 )
};
\addlegendentry{[less than $p$]}

\nextgroupplot[title={OLMo-2-32B-instruct}, xticklabels={}, yticklabels={}, ylabel={}]
\addplot+[mark=square*, orange, line width=2pt] coordinates {
  ($\frac{1}{10}$, 87.6 ) ($\frac{2}{10}$,  90 ) ($\frac{3}{10}$, 90.4  ) ($\frac{4}{10}$, 89.6 ) ($\frac{5}{10}$,  92.2 ) ($\frac{6}{10}$,  91 ) ($\frac{7}{10}$, 92.8  ) ($\frac{8}{10}$, 91.4  ) ($\frac{9}{10}$, 93.4  )
};
\addplot+[mark=*, teal, line width=1pt] coordinates {
  ($\frac{1}{10}$, 72 ) ($\frac{2}{10}$, 67.2  ) ($\frac{3}{10}$, 54.4  ) ($\frac{4}{10}$, 48 ) ($\frac{5}{10}$,  49.8 ) ($\frac{6}{10}$, 44  ) ($\frac{7}{10}$, 46.4  ) ($\frac{8}{10}$, 49  ) ($\frac{9}{10}$, 61.4  )
};

\nextgroupplot[title={K2-65B}, xticklabels={}, yticklabels={}, ylabel={}]
\addplot+[mark=square*, orange, line width=2pt] coordinates {
  ($\frac{1}{10}$, 93.4 ) ($\frac{2}{10}$,  92.2 ) ($\frac{3}{10}$,  89.8 ) ($\frac{4}{10}$, 92 ) ($\frac{5}{10}$, 92.6  ) ($\frac{6}{10}$,  91.4 ) ($\frac{7}{10}$, 91.4  ) ($\frac{8}{10}$, 93.8  ) ($\frac{9}{10}$, 92.8  )
};
\addplot+[mark=*, teal, line width=1pt] coordinates {
  ($\frac{1}{10}$, 89 ) ($\frac{2}{10}$,  86 ) ($\frac{3}{10}$, 82.8  ) ($\frac{4}{10}$, 79.8 ) ($\frac{5}{10}$,  79.2 ) ($\frac{6}{10}$, 77.2  ) ($\frac{7}{10}$, 85  ) ($\frac{8}{10}$,  84 ) ($\frac{9}{10}$, 90.8  )
};

\nextgroupplot[title={ (with explicit semantics)}, xlabel={$p$}]
\addplot+[mark=square*, orange, line width=2pt] coordinates {
  ($\frac{1}{10}$, 88.4 ) ($\frac{2}{10}$, 85.4  ) ($\frac{3}{10}$,  84 ) ($\frac{4}{10}$, 81 ) ($\frac{5}{10}$,  84.8 ) ($\frac{6}{10}$, 82.4  ) ($\frac{7}{10}$,  80.2 ) ($\frac{8}{10}$, 79.6  ) ($\frac{9}{10}$, 81  )
};
\addplot+[mark=*, teal, line width=1pt] coordinates {
  ($\frac{1}{10}$, 86.2 ) ($\frac{2}{10}$, 82  ) ($\frac{3}{10}$, 82.6  ) ($\frac{4}{10}$, 79.8 ) ($\frac{5}{10}$, 79.8  ) ($\frac{6}{10}$, 77.4  ) ($\frac{7}{10}$, 74  ) ($\frac{8}{10}$,  75.4 ) ($\frac{9}{10}$,  72 )
};

\nextgroupplot[title={ (with explicit semantics)},   yticklabels={},  ylabel={}]
\addplot+[mark=square*, orange, line width=2pt] coordinates {
  ($\frac{1}{10}$, 95.4 ) ($\frac{2}{10}$,  92 ) ($\frac{3}{10}$,  92.6 ) ($\frac{4}{10}$, 90.6 ) ($\frac{5}{10}$, 91.6  ) ($\frac{6}{10}$, 93.2  ) ($\frac{7}{10}$, 92.8  ) ($\frac{8}{10}$, 91.6  ) ($\frac{9}{10}$,  91.4 )
};
\addplot+[mark=*, teal, line width=1pt] coordinates {
  ($\frac{1}{10}$, 94.6 ) ($\frac{2}{10}$, 91.2  ) ($\frac{3}{10}$, 92.4  ) ($\frac{4}{10}$, 88.4  ) ($\frac{5}{10}$,  88 ) ($\frac{6}{10}$,  88.6 ) ($\frac{7}{10}$, 87.4  ) ($\frac{8}{10}$,  90.2 ) ($\frac{9}{10}$, 89.8  )
};

\nextgroupplot[title={ (with explicit semantics)},   yticklabels={}, ylabel={}]
\addplot+[mark=square*, orange, line width=2pt] coordinates {
  ($\frac{1}{10}$, 95.2 ) ($\frac{2}{10}$, 92  ) ($\frac{3}{10}$,  94 ) ($\frac{4}{10}$, 93 ) ($\frac{5}{10}$,  94.2 ) ($\frac{6}{10}$, 93.8  ) ($\frac{7}{10}$,  94.6 ) ($\frac{8}{10}$, 91.6  ) ($\frac{9}{10}$,  93.8 )
};
\addplot+[mark=*, teal, line width=1pt] coordinates {
  ($\frac{1}{10}$, 93.6 ) ($\frac{2}{10}$, 91.4  ) ($\frac{3}{10}$,  91.4 ) ($\frac{4}{10}$, 91.6 ) ($\frac{5}{10}$,  92.2 ) ($\frac{6}{10}$, 93.4  ) ($\frac{7}{10}$,  93.6 ) ($\frac{8}{10}$,  92.2 ) ($\frac{9}{10}$, 91  )
};

\end{groupplot}
\end{tikzpicture} }

    \caption{In-context concept learning helps uncover bias in monotonicity that is less noticeable in standard evaluation methods. In experiments with OLMo-2 and K2, models tend to have higher accuracies with upward monotone concepts during concept learning experiments (top). However, this bias is less noticeable in explicit semantics experiments (bottom).}
    
 \label{plot-2}

\end{figure}
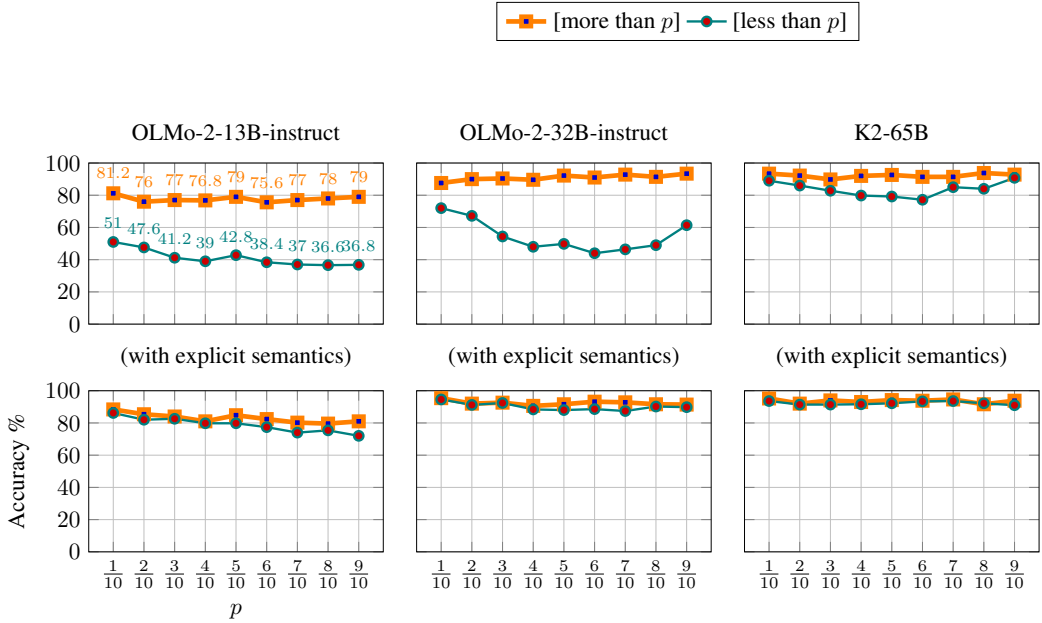

For each model, we run two sets of experiments -- one with concept learning and the other with explicit semantics. In figure \ref{plot-2}, the two charts in the first vertical column show the results for OLMo-2-13B-instruct. We see that the accuracy differences between concepts [more than $p$] and [less than $p$] are much greater in the concept learning set (top chart) when compared to the explicit semantics set (bottom chart). Similar patterns can be observed for OLMo-2-32B-instruct, represented by the two charts in the second vertical column. In the third column, we see that K2-65B\footnote{checkpoint 300 (out of 380) was used in this experiment} also has a bias toward upward monotone concepts, although the difference in accuracy between the two classes of concepts is smaller when compared to OLMo-2. This suggests that K2-65B may have a weaker bias toward upward monotonicity.

\section{Discussion}

The results with OLMo-2 and K2 models show that some LLMs consistently achieve lower accuracies for downward monotone quantifiers when learning new concepts.  The precise reason for this phenomenon is not yet known, but we have developed the following hypothesis:

\begin{itemize}
    \item \citep{more-less-01, more-less-02} have shown that downward monotone quantifiers can be expressed as the \textit{negation} of their upward monotone counterparts. They showed that humans generally require more processing time for downward monotone quantifiers (compared to upward monotone ones) in quantifier verification tasks, and further hypothesized that the hidden negation operation contributed to the increased complexity of downward monotone quantifiers.

    \item \citet{minimization-boolean-complexity-incontext} showed that LLMs can be biased toward logically simpler concepts when performing concept learning tasks. 

\end{itemize}

Based on the two observations above, we hypothesize that downward monotone concepts can be considered more logically complex than upward monotone ones in LLM concept learning because of the hidden negation operator. And similar to humans, certain LLMs perform worse on downward monotone concepts since they are biased toward logically simpler concepts, such as their upward monotone counterparts.

To summarize, we make the following contributions in this paper:

\begin{itemize}
    \item We show that LLMs can have an implicit bias toward mathematical concepts that are upward monotone. 
    \item We demonstrate that concept learning can be a new tool to uncover such bias.
    \item Lastly, we compare LLMs' performance with human performance on similar mathematical concepts, and find that LLMs have human-like biases in mathematical concept learning.
\end{itemize}

We hope this work will inspire new inquiries into uncovering hidden biases and studying how training data can enable certain biased behaviors in large language models. 
 
\bibliography{neurips_2025_dbm_findings}

@inproceedings{more-less-01,
  title={Measuring the cognitive cost of downward monotonicity by controlling for negative polarity},
  author={Galit Agmon and Yonatan Loewenstein and Yosef Grodzinsky},
  year={2019},
booktitle={Glossa: a journal of general linguistics 4(1): 36},
  url={https://www.glossa-journal.org/article/id/5143/}
}

@article{more-less-02,
  title={Negative sentences exhibit a sustained effect in delayed verification tasks.},
  author={Galit Agmon and Yonatan Loewenstein and Yosef Grodzinsky},
  journal={Journal of experimental psychology. Learning, memory, and cognition},
  year={2022},
  volume={48 1},
  pages={
          122-141
        },
  url={https://pubmed.ncbi.nlm.nih.gov/35254842/}
}

@incollection{goodmanConceptsProbabilisticLanguage2015,
  title = {Concepts in a {{Probabilistic Language}} of {{Thought}}},
  booktitle = {The {{Conceptual Mind}}},
  author = {Goodman, Noah D. and Tenenbaum, Joshua B. and Gerstenberg, Tobias},
  editor = {Margolis, Eric and Laurence, Stephen},
  date = {2015-05-08},
year = {2015},
  pages = {623--654},
  publisher = {The MIT Press},
  doi = {10.7551/mitpress/9383.003.0035},
  url = {https://direct.mit.edu/books/book/3430/chapter/115711/Concepts-in-a-Probabilistic-Language-of-Thought},
  urldate = {2024-09-20},
  isbn = {978-0-262-32686-5},
  langid = {english}
}

@article{piantadosiLogicalPrimitivesThought2016,
  title = {The Logical Primitives of Thought: {{Empirical}} Foundations for Compositional Cognitive Models},
  author = {Piantadosi, Steven T and Tenenbaum, Joshua B and Goodman, Noah D},
  year = {2016},
  journal = {Psychological Review},
  volume = {123},
  number = {4},
  pages = {392--424},
  doi = {10.1037/a0039980}
}

@article{feldmanMinimizationBooleanComplexity2000,
  title = {Minimization of {{Boolean Complexity}} in {{Human Concept Learning}}},
  author = {Feldman, Jacob},
  year = {2000},
  journal = {Nature},
  volume = {407},
  pages = {630--633},
  doi = {10.1038/35036586}
}

@inproceedings{min-etal-2022-rethinking,
    title = "Rethinking the Role of Demonstrations: What Makes In-Context Learning Work?",
    author = "Min, Sewon  and
      Lyu, Xinxi  and
      Holtzman, Ari  and
      Artetxe, Mikel  and
      Lewis, Mike  and
      Hajishirzi, Hannaneh  and
      Zettlemoyer, Luke",
    editor = "Goldberg, Yoav  and
      Kozareva, Zornitsa  and
      Zhang, Yue",
    booktitle = "Proceedings of the 2022 Conference on Empirical Methods in Natural Language Processing",
    month = dec,
    year = "2022",
    address = "Abu Dhabi, United Arab Emirates",
    publisher = "Association for Computational Linguistics",
    url = "https://aclanthology.org/2022.emnlp-main.759",
    doi = "10.18653/v1/2022.emnlp-main.759",
    pages = "11048--11064",
}

@misc{minimization-boolean-complexity-incontext,
      title={Minimization of Boolean Complexity in In-Context Concept Learning}, 
      author={Leroy Z. Wang and R. Thomas McCoy and Shane Steinert-Threlkeld},
      year={2024},
      eprint={2412.02823},
      archivePrefix={arXiv},
      primaryClass={cs.CL},
      url={https://arxiv.org/abs/2412.02823}, 
}

@article{implicit-bias-pnas,
author = {Xuechunzi Bai  and Angelina Wang  and Ilia Sucholutsky  and Thomas L. Griffiths },
title = {Explicitly unbiased large language models still form biased associations},
journal = {Proceedings of the National Academy of Sciences},
volume = {122},
number = {8},
pages = {e2416228122},
year = {2025},
doi = {10.1073/pnas.2416228122},
URL = {https://www.pnas.org/doi/abs/10.1073/pnas.2416228122},
eprint = {https://www.pnas.org/doi/pdf/10.1073/pnas.2416228122} }

@inproceedings{monotone-npi-license,
    title = "Language Models Use Monotonicity to Assess {NPI} Licensing",
    author = "Jumelet, Jaap  and
      Denic, Milica  and
      Szymanik, Jakub  and
      Hupkes, Dieuwke  and
      Steinert-Threlkeld, Shane",
    editor = "Zong, Chengqing  and
      Xia, Fei  and
      Li, Wenjie  and
      Navigli, Roberto",
    booktitle = "Findings of the Association for Computational Linguistics: ACL-IJCNLP 2021",
    month = aug,
    year = "2021",
    address = "Online",
    publisher = "Association for Computational Linguistics",
    url = "https://aclanthology.org/2021.findings-acl.439/",
    doi = "10.18653/v1/2021.findings-acl.439",
    pages = "4958--4969"
}

@inproceedings{bias-01-wan-etal-2023-kelly,
    title = "``Kelly is a Warm Person, Joseph is a Role Model'': Gender Biases in {LLM}-Generated Reference Letters",
    author = "Wan, Yixin  and
      Pu, George  and
      Sun, Jiao  and
      Garimella, Aparna  and
      Chang, Kai-Wei  and
      Peng, Nanyun",
    editor = "Bouamor, Houda  and
      Pino, Juan  and
      Bali, Kalika",
    booktitle = "Findings of the Association for Computational Linguistics: EMNLP 2023",
    month = dec,
    year = "2023",
    address = "Singapore",
    publisher = "Association for Computational Linguistics",
    url = "https://aclanthology.org/2023.findings-emnlp.243/",
    doi = "10.18653/v1/2023.findings-emnlp.243",
    pages = "3730--3748"
}

@inproceedings{bias-02-shirafuji-etal-2025-bias,
    title = "Bias Vector: Mitigating Biases in Language Models with Task Arithmetic Approach",
    author = "Shirafuji, Daiki  and
      Takenaka, Makoto  and
      Taguchi, Shinya",
    editor = "Rambow, Owen  and
      Wanner, Leo  and
      Apidianaki, Marianna  and
      Al-Khalifa, Hend  and
      Eugenio, Barbara Di  and
      Schockaert, Steven",
    booktitle = "Proceedings of the 31st International Conference on Computational Linguistics",
    month = jan,
    year = "2025",
    address = "Abu Dhabi, UAE",
    publisher = "Association for Computational Linguistics",
    url = "https://aclanthology.org/2025.coling-main.190/",
    pages = "2799--2813"
}

@article{bias-03-Hofmann2024AIGC,
  title={AI generates covertly racist decisions about people based on their dialect},
  author={Valentin Hofmann and Pratyusha Kalluri and Dan Jurafsky and Sharese King},
  journal={Nature},
  year={2024},
  volume={633},
  pages={147 - 154},
  url={https://api.semanticscholar.org/CorpusID:272214842}
}

@article{icard-iii-moss-2014-monotonicity,
    title = "Recent Progress on Monotonicity",
    author = "Icard III, Thomas F.  and
      Moss, Lawrence S.",
    journal = "Linguistic Issues in Language Technology",
    volume = "9",
    year = "2014",
    publisher = "CSLI Publications",
    url = "https://aclanthology.org/2014.lilt-9.7/"
}

@inproceedings{implicit-bias-02-tan-lee-2025-unmasking,
    title = "Unmasking Implicit Bias: Evaluating Persona-Prompted {LLM} Responses in Power-Disparate Social Scenarios",
    author = "Tan, Bryan Chen Zhengyu  and
      Lee, Roy Ka-Wei",
    editor = "Chiruzzo, Luis  and
      Ritter, Alan  and
      Wang, Lu",
    booktitle = "Proceedings of the 2025 Conference of the Nations of the Americas Chapter of the Association for Computational Linguistics: Human Language Technologies (Volume 1: Long Papers)",
    month = apr,
    year = "2025",
    address = "Albuquerque, New Mexico",
    publisher = "Association for Computational Linguistics",
    url = "https://aclanthology.org/2025.naacl-long.50/",
    doi = "10.18653/v1/2025.naacl-long.50",
    pages = "1075--1108",
    ISBN = "979-8-89176-189-6"
}

@article{quantifiers-Barwise1981GeneralizedQA,
  title={Generalized quantifiers and natural language},
  author={Jon Barwise and R. Cooper},
  journal={Linguistics and Philosophy},
  year={1981},
  volume={4},
  pages={159-219},
  url={https://api.semanticscholar.org/CorpusID:62189594}
}

@article{monotone-quant-01-Carcassi2021MonotoneQE,
  title={Monotone Quantifiers Emerge via Iterated Learning},
  author={Fausto Carcassi and Shane Steinert-Threlkeld and Jakub Szymanik},
  journal={Cognitive Science},
  year={2021},
  volume={45},
  url={https://api.semanticscholar.org/CorpusID:236976564}
}

@article{monotonicity-processing-load-Geurts2005,
  title={Monotonicity and Processing Load},
  author={Bart Geurts and Frans van der Slik},
  journal={J. Semant.},
  year={2005},
  volume={22},
  pages={97-117},
  url={https://api.semanticscholar.org/CorpusID:16068356}
}

@article{in-context-learn-01-Wei2022EmergentAO,
  title={Emergent Abilities of Large Language Models},
  author={Jason Wei and Yi Tay and Rishi Bommasani and Colin Raffel and Barret Zoph and Sebastian Borgeaud and Dani Yogatama and Maarten Bosma and Denny Zhou and Donald Metzler and Ed H. Chi and Tatsunori Hashimoto and Oriol Vinyals and Percy Liang and Jeff Dean and William Fedus},
  journal={ArXiv},
  year={2022},
  volume={abs/2206.07682},
  url={https://api.semanticscholar.org/CorpusID:249674500}
}

@article{in-context-learn-02-Kojima2022LargeLM,
  title={Large Language Models are Zero-Shot Reasoners},
  author={Takeshi Kojima and Shixiang Shane Gu and Machel Reid and Yutaka Matsuo and Yusuke Iwasawa},
  journal={ArXiv},
  year={2022},
  volume={abs/2205.11916},
  url={https://api.semanticscholar.org/CorpusID:249017743}
}

@article{OLMo20242O2,
  title={2 OLMo 2 Furious},
  author={Team OLMo and Pete Walsh and Luca Soldaini and Dirk Groeneveld and Kyle Lo and Shane Arora and Akshita Bhagia and Yuling Gu and Shengyi Huang and Matt Jordan and Nathan Lambert and Dustin Schwenk and Oyvind Tafjord and Taira Anderson and David Atkinson and Faeze Brahman and Christopher Clark and Pradeep Dasigi and Nouha Dziri and Michal Guerquin and Hamish Ivison and Pang Wei Koh and Jiacheng Liu and Saumya Malik and William Merrill and Lester James Validad Miranda and Jacob Daniel Morrison and Tyler C. Murray and Crystal Nam and Valentina Pyatkin and Aman Rangapur and Michael Schmitz and Sam Skjonsberg and David Wadden and Christopher Wilhelm and Michael Wilson and Luke S. Zettlemoyer and Ali Farhadi and Noah A. Smith and Hanna Hajishirzi},
  journal={ArXiv},
  year={2024},
  volume={abs/2501.00656},
  url={https://api.semanticscholar.org/CorpusID:275213098}
}

@article{k2-llm,
  title={LLM360 K2: Building a 65B 360-Open-Source Large Language Model from Scratch},
  author={Zhengzhong Liu and Bowen Tan and Hongyi Wang and Willie Neiswanger and Tianhua Tao and Haonan Li and Fajri Koto and Yuqi Wang and Suqi Sun and Omkar Pangarkar and Richard Fan and Yi Gu and Victor Miller and Liqun Ma and Liping Tang and Nikhil Ranjan and Yonghao Zhuang and Guowei He and Renxi Wang and Ming Deng and Robin Algayres and Yuanzhi Li and Zhiqiang Shen and Preslav Nakov and Eric Xing},
  journal={ArXiv},
  year={2025},
  volume={abs/2501.07124},
  url={https://api.semanticscholar.org/CorpusID:275471059}
}

@article{Touvron2023Llama2O,
  title={Llama 2: Open Foundation and Fine-Tuned Chat Models},
  author={Hugo Touvron and Louis Martin and Kevin R. Stone and Peter Albert and Amjad Almahairi and Yasmine Babaei and Niko-lay Bashlykov and Soumya Batra and Prajjwal Bhargava and Shruti Bhosale and Daniel M. Bikel and Lukas Blecher and Cris-tian Cant{\'o}n Ferrer and Moya Chen and Guillem Cucurull and David Esiobu and Jude Fernandes and Jeremy Fu and Wenyin Fu and Brian Fuller and Cynthia Gao and Vedanuj Goswami and Naman Goyal and Anthony S. Hartshorn and Saghar Hosseini and Rui Hou and Hakan Inan and Marcin Kardas and Viktor Kerkez and Madian Khabsa and Isabel M. Kloumann and Artem Korenev and Punit Singh Koura and Marie-Anne Lachaux and Thibaut Lavril and Jenya Lee and Diana Liskovich and Yinghai Lu and Yuning Mao and Xavier Martinet and Todor Mihaylov and Pushkar Mishra and Igor Molybog and Yixin Nie and Andrew Poulton and Jeremy Reizenstein and Rashi Rungta and Kalyan Saladi and Alan Schelten and Ruan Silva and Eric Michael Smith and R. Subramanian and Xia Tan and Binh Tang and Ross Taylor and Adina Williams and Jian Xiang Kuan and Puxin Xu and Zhengxu Yan and Iliyan Zarov and Yuchen Zhang and Angela Fan and Melissa Hall Melanie Kambadur and Sharan Narang and Aur'elien Rodriguez and Robert Stojnic and Sergey Edunov and Thomas Scialom},
  journal={ArXiv},
  year={2023},
  volume={abs/2307.09288},
  url={https://api.semanticscholar.org/CorpusID:259950998}
}


\appendix

\section{Limitations}
Only a limited set of LLMs were tested; we hope to apply the concept learning methodology to study a larger set of models with greater variety as future work.

The set of upward/downward monotone concepts is relatively small. The next step can be to study this phenomenon with more complex concepts with varying monotonicity.

\section{Additional concept learning results}\label{additional-concepts}

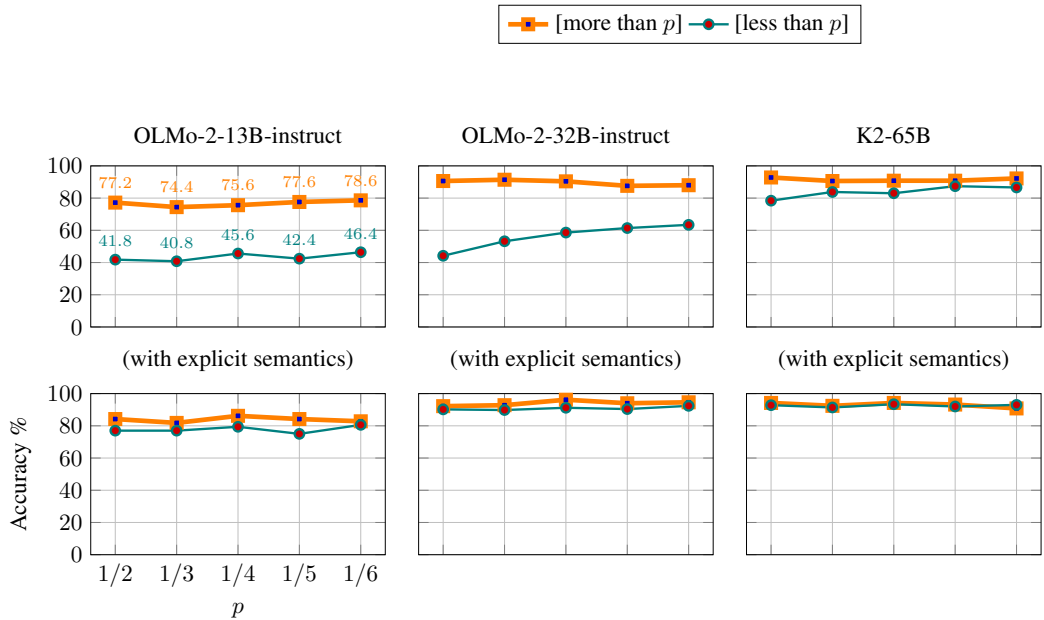
\begin{figure}[H]
    \centering
\resizebox{\textwidth}{!}{

\begin{tikzpicture} 
\begin{groupplot}[
  group style={
    group size=3 by 2,
    horizontal sep=0.5cm,
    vertical sep=1cm
  },
  width=6cm,
  height=4cm,
  xlabel={ },
  ylabel={Accuracy \%},
  ymin=0, ymax=100,
  legend style={at={(2, 2)},anchor=north,legend columns=-1},
  symbolic x coords={$1/2$,$1/3$,$1/4$,$1/5$,$1/6$},
  xtick=data,
  grid=both,
]

\nextgroupplot[title={OLMo-2-13B-instruct}, xticklabels={}, ylabel={}]
\addplot+[mark=square*, orange, line width=2pt,   
nodes near coords,
  point meta=y,
  every node near coord/.append style={font=\scriptsize, yshift=2pt}] coordinates {
  ($1/2$,77.2) ($1/3$,74.4) ($1/4$,75.6) ($1/5$,77.6) ($1/6$,78.6)
};
\addlegendentry{[more than $p$]}
\addplot+[mark=*, teal, line width=1pt,
nodes near coords,
  point meta=y,
  every node near coord/.append style={font=\scriptsize, yshift=2pt}] coordinates {
  ($1/2$,41.8) ($1/3$,40.8) ($1/4$,45.6) ($1/5$,42.4) ($1/6$,46.4)
};
\addlegendentry{[less than $p$]}

\nextgroupplot[title={OLMo-2-32B-instruct}, xticklabels={}, yticklabels={}, ylabel={}]
\addplot+[mark=square*, orange, line width=2pt] coordinates {
  ($1/2$,90.6) ($1/3$,91.4) ($1/4$,90.4) ($1/5$,87.6) ($1/6$,88)
};
\addplot+[mark=*, teal, line width=1pt] coordinates {
  ($1/2$,44.2) ($1/3$,53.2) ($1/4$,58.6) ($1/5$,61.4) ($1/6$,63.4)
};

\nextgroupplot[title={K2-65B}, xticklabels={}, yticklabels={}, ylabel={}]
\addplot+[mark=square*, orange, line width=2pt] coordinates {
  ($1/2$, 92.8 ) ($1/3$, 90.6 ) ($1/4$, 90.8 ) ($1/5$, 90.8 ) ($1/6$, 92.2)
};
\addplot+[mark=*, teal, line width=1pt] coordinates {
    ($1/2$, 78.4 ) ($1/3$, 83.8 ) ($1/4$, 83.0 ) ($1/5$, 87.4 ) ($1/6$, 86.6 )
};

\nextgroupplot[title={ (with explicit semantics)}, xlabel={$p$}]
\addplot+[mark=square*, orange, line width=2pt] coordinates {
  ($1/2$, 84.2) ($1/3$, 81.8) ($1/4$,86.2) ($1/5$,84.2) ($1/6$,82.8)
};
\addplot+[mark=*, teal, line width=1pt] coordinates {
  ($1/2$,77) ($1/3$,77 ) ($1/4$,79.4) ($1/5$,75) ($1/6$,80.6)
};

\nextgroupplot[title={ (with explicit semantics)}, xticklabels={}, yticklabels={}, ylabel={}]
\addplot+[mark=square*, orange, line width=2pt] coordinates {
  ($1/2$,92.2) ($1/3$,92.8) ($1/4$,96.2) ($1/5$,94) ($1/6$,94.6)
};
\addplot+[mark=*, teal, line width=1pt] coordinates {
  ($1/2$,90.2) ($1/3$,89.8) ($1/4$,91.2) ($1/5$,90.4) ($1/6$,92.4)
};

\nextgroupplot[title={ (with explicit semantics)}, xticklabels={}, yticklabels={}, ylabel={}]
\addplot+[mark=square*, orange, line width=2pt] coordinates {
  ($1/2$, 94.2 ) ($1/3$, 92.4) ($1/4$, 94.2 ) ($1/5$, 93.2 ) ($1/6$, 90.8 )
};
\addplot+[mark=*, teal, line width=1pt] coordinates {
  ($1/2$, 92.8) ($1/3$, 91.4 ) ($1/4$, 93.4 ) ($1/5$, 92  ) ($1/6$, 93 )
};

\end{groupplot}
\end{tikzpicture} }

    \caption{In-context concept learning helps uncover bias in monotonicity that is less noticeable in standard evaluation methods. In experiments with OLMo-2 and K2, models tend to have higher accuracies with upward monotone concepts during concept learning experiments (top). However, this bias is less noticeable in explicit semantics experiments (bottom).}
    
 \label{fig:main-plot2}
    
\end{figure}

\begin{table}[H]
\centering
\caption{Accuracy values}
\renewcommand{\arraystretch}{1.2} 
\begin{tabularx}{\textwidth}{l *{5}{c}}
\toprule
Model & $1/2$ & $1/3$ & $1/4$ & $1/5$ & $1/6$ \\
\midrule

\multicolumn{6}{l}{\textbf{OLMo-2-13B-instruct}} \\
\quad [more than p] & 77.2 & 74.4 & 75.6 & 77.6 & 78.6 \\
\quad [less than p] & 41.8 & 40.8 & 45.6 & 42.4 & 46.4 \\

\addlinespace
\multicolumn{6}{l}{\textbf{OLMo-2-13B-instruct (explicit semantics)}} \\
\quad [more than p] & 84.2 & 81.8 & 86.2 & 84.2 & 82.8 \\
\quad [less than p] & 77.0 & 77.4 & 79.4 & 75.0 & 80.6 \\

\addlinespace
\multicolumn{6}{l}{\textbf{OLMo-2-32B-instruct}} \\
\quad [more than p] & 90.6 & 91.4 & 90.4 & 87.6 & 88.0 \\
\quad [less than p] & 44.2 & 53.2 & 58.6 & 61.4 & 63.4 \\

\addlinespace
\multicolumn{6}{l}{\textbf{OLMo-2-32B-instruct (explicit semantics)}} \\
\quad [more than p] & 92.2 & 92.8 & 96.2 & 94.0 & 94.6 \\
\quad [less than p] & 90.2 & 89.8 & 91.2 & 90.4 & 92.4 \\

\addlinespace
\multicolumn{6}{l}{\textbf{K2-65B}} \\
\quad [more than p] & 92.8 & 90.6 & 90.8 & 90.8 & 92.2 \\
\quad [less than p] & 78.4 & 83.8 & 83.0 & 87.4 & 86.6 \\

\addlinespace
\multicolumn{6}{l}{\textbf{K2-65B (explicit semantics)}} \\
\quad [more than p] & 94.2 & 92.4 & 94.2 & 93.2 & 90.8 \\
\quad [less than p] & 92.8  & 91.4 & 93.4 & 92.0 & 93.0 \\

\bottomrule
\end{tabularx}
\end{table}

\subsection{Cardinal concepts}

\begin{figure}[h!]
\centering
\begin{tikzpicture}
\begin{axis}[
title={OLMo-2-32B-instruct},
    width=11cm,
    height=7cm,
    xlabel={$c$},
    ylabel={Accuracy \%},
    ylabel style={rotate=-90},
    ymin=0, ymax=100,
    xmin=3, xmax=30,
    xtick={3,5,7,9,11,13,15,17,19,21,23,25,27,29},
    grid=both,
    legend style={
        at={(1, 1.1)},
        anchor=south,
        legend columns=-1,
        font=\small 
    },
    tick label style={font=\small},
    label style={font=\small},
]

\addplot[
    thick,
    color=teal,
    mark=*,
    mark options={fill=red, scale=1},
    line width=1pt
] coordinates {
(3,87.0)
(4,84.6)
(5,79.8)
(6,75.2)
(7,75.2)
(8,75.2)
(9,74.8)
(10,69.0)
(11,68.0)
(12,68.6)
(13,66.2)
(14,66.2)
(15,61.8)
(16,63.8)
(17,63.4)
(18,59.4)
(19,56.4)
(20,60.0)
(21,59.0)
(22,55.6)
(23,57.6)
(24,57.0)
(25,60.0)
(26,56.2)
(27,60.0)
(28,58.2)
(29,61.2)
(30,56.8)
};
 \addlegendentry{[less than $c$]}

\addplot[
    thick,
    color=orange,
    mark=square*,
    mark options={fill=orange!50!yellow, scale=1},
    line width=2pt
] coordinates {
(3,85.8)
(4,85.4)
(5,83.2)
(6,84.0)
(7,83.0)
(8,85.2)
(9,84.6)
(10,82.2)
(11,85.8)
(12,80.8)
(13,86.8)
(14,83.8)
(15,83.8)
(16,82.4)
(17,83.8)
(18,82.2)
(19,83.0)
(20,83.6)
(21,82.8)
(22,84.2)
(23,84.4)
(24,85.2)
(25,82.8)
(26,83.8)
(27,83.8)
(28,83.4)
(29,84.0)
(30,84.6)
};
 \addlegendentry{[more than $c$]}

\end{axis}
\end{tikzpicture}

\caption{OLMo-2-32B concept learning results with cardinal concepts. The bias toward upward monotonicity becomes stronger as $c$ increases. $c \in [3, 30] $.}

\end{figure}
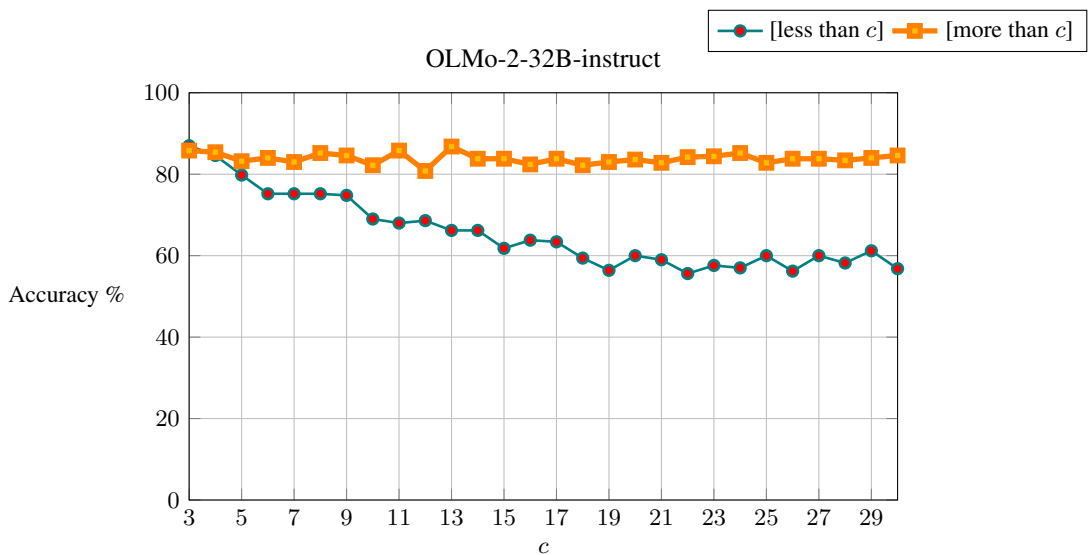

\section{Example prompt} \label{example-prompt}

\begin{table}[H]
\centering

\begin{tabular}{ l|l } 
\hline
\textbf{Prompt} & \textbf{Label} \\
\hline

There are 76 fields. Alice has 65 of the 76 fields. & \\
Does Alice have the desired quantity of the fields? No. & \\
There are 74 houses. Alice has 17 of the 74 houses. & \\
Does Alice have the desired quantity of the houses? Yes.& \\
There are 51 rolls. Alice has 18 of the 51 rolls. & \\
Does Alice have the desired quantity of the rolls? Yes.& \\ 
There are 79 beds. Alice has 52 of the 79 beds. & \\
Does Alice have the desired quantity of the beds? No.& \\ 
There are 89 boards. Alice has 46 of the 89 boards. & \\
Does Alice have the desired quantity of the boards? No.& \\
There are 82 books. Alice has 42 of the 82 books. & \\
Does Alice have the desired quantity of the books? No.& \\ 
There are 96 texts. Alice has 6 of the 96 texts.& \\
Does Alice have the desired quantity of the texts? Yes.& \\ 
There are 94 cars. Alice has 49 of the 94 cars. & \\
Does Alice have the desired quantity of the cars? No.& \\ 
There are 81 forces. Alice has 65 of the 81 forces.& \\
Does Alice have the desired quantity of the forces? No.& \\ 
There are 81 whiles. Alice has 75 of the 81 whiles. & \\
Does Alice have the desired quantity of the whiles? No.& \\ 
There are 90 endings. Alice has 52 of the 90 endings. & \\
Does Alice have the desired quantity of the endings? No.& \\ 
There are 79 lips. Alice has 3 of the 79 lips. & \\
Does Alice have the desired quantity of the lips? Yes.& \\ 
There are 17 men. Alice has 4 of the 17 men. & \\
Does Alice have the desired quantity of the men? Yes.& \\ 
There are 42 states. Alice has 36 of the 42 states. & \\
Does Alice have the desired quantity of the states? No.& \\ 
There are 73 rooms. Alice has 29 of the 73 rooms. & \\
Does Alice have the desired quantity of the rooms? Yes.& \\
 There are 64 waters. Alice has 11 of the 64 waters. & \\
 Does Alice have the desired quantity of the waters? Yes.& \\ 
There are 50 friends. Alice has 8 of the 50 friends. & \\
Does Alice have the desired quantity of the friends? Yes.& \\ 
There are 28 rooms. Alice has 8 of the 28 rooms. & \\
Does Alice have the desired quantity of the rooms? Yes.& \\ 
There are 56 bits. Alice has 56 of the 56 bits.& \\
Does Alice have the desired quantity of the bits? No.& \\ 
There are 96 apps. Alice has 9 of the 96 apps. & \\
Does Alice have the desired quantity of the apps? Yes.& \\ 
There are 95 boxes. Alice has 7 of the 95 boxes. & \\
Does Alice have the desired quantity of the boxes?& \\

 & Yes \\ 

\hline

\end{tabular}

\caption{An example of prompts in the dataset. The concept in this example is ``less than half''. In an \textit{explicit semantics} experiment, the prompt will have a similar structure, but the phrase ``the desired quantity'' will be replaced by ``less than $1/2$''.}
\label{table-prompt}
\end{table}

\end{document}